\documentclass{article}

\usepackage[preprint]{neurips_2023}
\usepackage{amsmath,amssymb,amsfonts}
\usepackage{graphicx}
\usepackage{subfig}
\usepackage[utf8]{inputenc} % allow utf-8 input
\usepackage[T1]{fontenc}    % use 8-bit T1 fonts
\usepackage{hyperref}       % hyperlinks
\usepackage{url}            % simple URL typesetting
\usepackage{booktabs}       % professional-quality tables
\usepackage{amsfonts}       % blackboard math symbols
\usepackage{nicefrac}       % compact symbols for 1/2, etc.
\usepackage{microtype}      % microtypography
\usepackage{xcolor}         % colors

\title{Skin Cancer Images Classification using 
Transfer Learning Techniques}

\author{%
  Md Sirajul Islam\\
  School of Computing and Informatics\\
  University of Louisiana at Lafayette\\
  Lafayette, LA 70503 \\
  \texttt{md-sirajul.islam1@louisiana.edu} \\
  \And
  Sanjeev Panta\\
  School of Computing and Informatics\\
  University of Louisiana at Lafayette\\
  Lafayette, LA 70503 \\
  \texttt{sanjeev.panta1@louisiana.edu} \\
}

\begin{document}

\maketitle

\begin{abstract}
  Skin cancer is one of the most common and deadliest types of cancer. Early diagnosis of skin cancer at a benign stage is critical to reducing cancer mortality. To detect skin cancer at an earlier stage an automated system is compulsory that can save the life of many patients. Many previous studies have addressed the problem of skin cancer diagnosis using various deep learning and transfer learning models. However, existing literature has limitations in its accuracy and time-consuming procedure. In this work, we applied five different pre-trained transfer learning approaches for binary classification of skin cancer detection at benign and malignant stages. To increase the accuracy of these models we fine-tune different layers and activation functions. We used a publicly available ISIC dataset to evaluate transfer learning approaches. For model stability, data augmentation techniques are applied to improve the randomness of the input dataset. These approaches are evaluated using different hyperparameters such as batch sizes, epochs, and optimizers. The experimental results show that the ResNet-50 model provides an accuracy of 0.935, F1-score of 0.86, and precision of 0.94. 
\end{abstract}

\section{Introduction}

Skin cancer is one of the most common types of cancer all over the world. The most common cause is excessive exposure of the skin to UV rays from the sun [1]. Fair-skinned, sun-sensitive people have a higher rate of UV exposure than dark-skinned people who are less sensitive to the sun [2]. There are various types of skin cancer. Among them, melanoma skin cancer is the most common and dangerous type of cancer. It involves an abnormal proliferation of pigment cells in the skin. It is a common life-threatening disease in humans. It brightens the skin and can quickly spread to other areas body [3]. Therefore, early detection of melanoma is essential. Invasive melanoma accounts for approximately 1\% of all skin cancer cases but contributes to most skin cancer deaths. The incidence of melanoma skin cancer has increased rapidly over the past 30 years. In 2021, it was estimated that 100,350 new cases of melanoma were diagnosed in the United States, ultimately resulting in about 6,850 deaths [4].

Early detection and prevention are the best way to control skin cancer [5]. To perform skin tests diagnostic facilities as well as medical professionals are required [6]. However, earlier detection may not always be possible due to the time-consuming manual diagnosis procedures. Furthermore, many poor people are unable to afford this costlier diagnosis. Perception of new or changing skin patches or growths, especially those that appear abnormal, should be assessed. New lesions or progressive changes in lesion appearance (size, color, or shape) should be evaluated by a doctor. That's why we need to develop deep learning models that outperform human diagnostics and produce less time-consuming and accurate results [7]. Deep learning models can classify malignant skin cancers such as melanoma and benign skin lesions.

Various research [8] have been carried out on skin cancer detection using CNNs to classify images into two types called melanoma and non-melanoma. However, this binary classification approach does not consider other types of skin cancers [9]. Therefore, many researchers have developed research focusing on the multi-class classification of skin cancer using CNN models, and have achieved high accuracy. However, the work is limited to the application of a single CNN model, and other methods such as data augmentation to increase the variety of images and help rotate, flip, or resize images are not considered. We use data augmentation techniques with CNN models to provide unbiased results and solve the problem of data imbalance [6]. Our work focuses on increasing the performance of existing approaches of binary image classification by using transfer learning and data augmentation. Our model aims to perform binary classification of images to identify benign and malignant skin lesions in the ISIC dataset with CNNs. 

In this work, we aim to classify skin cancer images that can assist in earlier skin cancer detection. Considering the complexity involved in the diagnosis of skin cancer images at an earlier stage, this work focuses on the following:

\begin{itemize}
\item A transfer learning-based paradigm for the classification of skin cancer into benign and malignant.
\item To provide stability to models and randomness of the input dataset we employ data augmentation techniques in the data pre-processing stage.
\item An experimental approach, changing different network hyperparameters to observe the corresponding impact – positive or negative on the performance of the network.
\item Reporting the network’s performance in terms of various performance metrics. 
\item Compare the performance of various transfer learning approaches.
\end{itemize}

The rest of the paper is organized as follows: Section 2 describes the related work in the classification of Melanoma
skin cancer. Section 3 describes the proposed approach for solving the skin cancer detection and classification of skin lesions. Section 4 provides the experimental results of the proposed transfer learning models. Section 5 presents the conclusion and findings from the results.

\section{Related Works}
Many new methods have emerged in recent years given the rise in the importance of the detection and classification of skin cancer from dermatoscopy skin lesions images. Computer-assisted diagnosis systems become fast and inexpensive alternatives for detecting various types of diseases. Recently, deep learning techniques have made a huge impact on various problems including medical image processing [10], [11], [12], [21]. It is challenging to perform research on medical data due to the unavailability of large and accurate datasets. Medical institutions do not like to share privacy-sensitive patient data. Despite of having such challenges many prior research is able to assist healthcare practitioners at a large scale. In the past few years,  several works aimed at skin cancer detection and classification using CNNs have been published. Although the goals of all these tasks are very similar, each task focuses on a specific type of problem. For example, [17] is proposing a two-class (malignant and benign) classifier.  Other works in [11] consider more complex 7-class  classification (melanoma, melanocytic nevus, basal cell carcinoma, actinic keratosis, benign keratosis, dermatofibroma, and vascular lesion). In the next few paragraphs, we will discuss about prior research those focuses on the classification of skin cancer images using deep learning techniques.

Milton et al. [11] proposed an approach to classify Melanoma and other skin lesion cancer images. The author conducted experiments by using the ISIC 2018 dataset on various deep learning models such as PNASNet-5-large, InceptionResNetV2, SENet154, and InceptionV4. They trained their models on different settings by changing parameters and learning rates, and optimizers. The experimental results displayed a class number for each type of skin lesion. Their experimental results provided the highest validation score of 76\% for the PNASNet-5-large model. This work showed good classification results for a variety of skin lesions but did not consider the transfer learning process. One of the effective techniques to reduce bias and oversampling in datasets is data augmentation. It is more efficient when the dataset size is small. It helps to augment the data by rotating, flipping, and cropping the images. Another way of achieving higher accuracy in the case of smaller datasets is transfer learning. Sometimes a small amount of data is not enough to train a deep learning model. It is therefore advantageous to use pre-trained models to overcome this problem.

Nugroho et al. [15] constructed a skin cancer detection system by using a customized CNN model. They used the HAM10000 skin cancer dataset. They achieved an accuracy of 78\%. Bassi et al. [17] introduced a transfer learning-based deep learning strategy on the HAM10000 dataset. They fine-tuned the VGG-16 model as well as resized HAM10000 dataset photos into 224×224. They attained 82.8\% accuracy from their model. Chaturvedi et al. [16] presented the detection of seven different types of skin lesions using MobileNet. They were able to achieve an accuracy of 83\%. Canziani et al. [18] applied two types of machine learning algorithms such as K-Means and Support Vector Machine. Their proposed approach provided an accuracy of 90\%. Moldovan et al. [19] exploit a transfer learning based technique. They made an accuracy of 85\% on the HAM10000 dataset.

Matsunaga et al. [20] classified three types of images such as melanoma, nevus, and seborrheic keratosis by using the ISIC dataset for skin lesion classification. They applied pre-trained deep learning models on ImageNet, ResNet-101, and InceptionV4 to carry out their experiments. They also used data augmentation for flipping, cropping, and zooming the images. The author normalized the input images to produce unbiased results for image classification. Our research is based on the above approaches. Our work's aim is to classify binary types of skin lesions as well as to improve model accuracy. For this purpose, we used different transfer learning techniques to train our CNN model with various combinations of hyperparameters to improve model performance. We applied five different transfer learning architectures such as ResNet-50, VGG16, MobileNet, InceptionV3, and DenseNet-169. A publicly available ISIC dataset was used for training and testing our models. We have used data augmentation techniques to improve classification results. For evaluating the performance of different transfer learning models we used some common performance matrices such as F1 score, precision, recall, and ROC-AUC. Our work provides an accuracy of 0.935, recall of 0.77, F1 score of 0.85, precision of 0.94, and ROC-AUC of 0.861.

\begin{table}[htbp]
\caption{Dataset Splitting}
\centering
\begin{tabular}{lllll}
\toprule
\text{Type} & \text{Total} & \text{Training} & \text{Testing} & \text{Validation}\\
\midrule
Benign&1800& 1534 & 186 & 80\\
Malignant&1497& 1284 & 144 & 69\\
Total&3297& 2818& 330&149\\
\bottomrule
\end{tabular}
\end{table}
\subsection{Dataset}
The dataset which is used in this research obtained from the Kaggle database [21] and consists of 3297 skin cancer images. There are 1800 benign and 1497 malignant RGB images. The dimensions of each image are (224 × 224 × 3) pixels. Fig. 1 shows two samples of benign and malignant skin cancer images from the dataset. The dataset consists of 3297 images out of which 1800 images are benign and 1497 images are malignant. We partition the dataset into training, testing, and validation. 10\% of benign and malignant images are used for the testing, and 5\% for the validation. The remaining images are used for training the model. Table 1 shows the number of images used for training, testing, and validation for both skin cancer classes. 
\section{Methodology}
\begin{figure*}
\centering
\subfloat[Benign images]
{\includegraphics[height=4cm, width=0.5\textwidth]{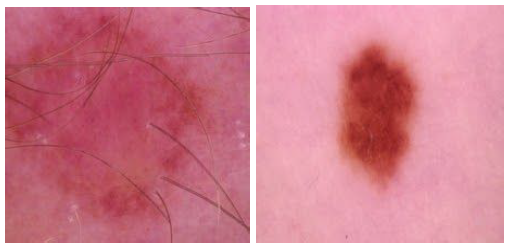}\label{fig:1}}
\subfloat[Malignant images]
{\includegraphics[height=4cm, width=0.5\textwidth]{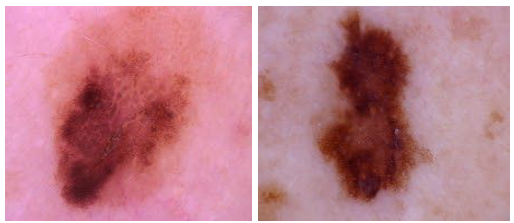}\label{fig:2}}
\caption{Illustration of the Benign and Malignant samples. }
\label{figure}
\end{figure*}
\subsection{Convolutional neural networks}
A Convolutional Neural Network (CNN) is a special type of artificial neural network (ANN) that is used to process pixel data. It is used for image classification, segmentation, and processing.  ConvNets [22] involve application of local feature detectors or filters over the whole image to measure the correspondence between individual image patches and signature patterns within the training set. Then, an aggregation or pooling function is applied to reduce the dimensionality of the feature space. A CNN consists of multiple layers such as the input layer, an output layer, and hidden layers (multiple convolutional layers, pooling layers, fully connected layers and normalization layers). The basic building blocks of a convolutional neural network are described below:

\begin{itemize}
\item \textbf{Input layer:} This is the first layer of a CNN. It receives input data from the exterior system. It acts as a provider for the whole CNN.
\item \textbf{Convolutional layer:} The main function of this layer is the extraction of features of the input layer. This process involves applying a 2D convolution of the input feature maps (i.e. image channels for the first layer) with a squared kernel (or filter).
\item \textbf{Activation function:} It is one of the important parts of a CNN that determines which node will fire or not. The ReLU (·) function is used to rectify non-linearities in the convolution output. This function helps us assign zero to all negative inputs and positive values remain.
\item \textbf{Pooling layer:} This stage allows to reduce the original large dimension of image representation through a subsampling strategy that supports local space invariances. This layer also controlled overfitting. It is used to reduce the memory of the network and make CNN faster. In this work, we used max-pooling. 
\item \textbf{Fully-connected layer:} This layer is typically applied in the top layer of a CNN architecture in order to capture complex relationships between high-level features. It uses a soft-max activation function to classify generated features of the input images into target classes. 
\item \textbf{Classification layer:} This final layer is a fully connected layer with one neuron per each of the two classes (benign or malignant). We used the soft-max activation function to predict the probability of a class as an output.
\end{itemize}

\subsection{Transfer Learning}
Transfer learning is a machine learning technique in which a model developed from one domain is reused in another domain. It is mainly useful in the case when we do not have enough training data. However, the required amount of data can be increased by applying data augmentation. One of the main reasons for using transfer learning due to the high similarity between the malignant and benign images. It will take a longer time to detect and classify images without transfer learning. Moreover, it becomes our first choice because of the effective detection of similar images. Generally, transfer learning models are trained on larger datasets. We use these pre-trained models and finetune a few layers to fit our dataset. In this work, we used five different transfer learning models. They are ResNet-50, MobileNet, InceptionV3, DenseNet-169, and InceptionResNetV2. We re-trained these models using our dataset. We analyze their predictions by applying these transfer learning techniques to the skin lesion dataset.

\textbf{ResNet-50:} ResNet is the shorter form of residual network. It could provide powerful performance that trains very large convolutional neural networks with many layers [22]. Parameter selection is dependent on the number of layers in the network. It performs batch normalization after each convolutional layer. It can reduce the issue of covariate shift and vanishing gradient descent problems. We re-trained the ResNet50 model on our dataset. We fine-tuned across all layers and replace the top layers with one average pooling and one fully connected layer to fit our dataset. Finally, we added a soft-max activation function that classifies images into two categories. To fit with this model, we resized all input images into (224 × 224 × 3) pixels. We used the Adam optimizer and the learning rate was 0.0001.

\textbf{InceptionV3:} It is the modified version of the GoogLeNet architecture [23]. The core idea of this model is to make this procedure more effective, and simple. This model performs as a multi-level feature extractor. We fine-tuned across all layers of Google’s InceptionV3 model. We replace the top layers with one average pooling and two fully connected layers. The size of the input images, learning rate, and optimizer remain the same as ResNet50. We re-train the model on our dataset and used the soft-max activation function for classification.

\textbf{InceptionResnetv2:} It used a residual version of Inception nets instead of the simple inception models [24]. We fine-tuned the model across all layers to retrain on our dataset. We replace the top layers with one average pooling and one fully connected layer. We have used the same input sizes, learning rates, optimizer, and activation functions for this model. 

\textbf{MobileNet:} The core part of this model is the depth-wise separable filters. Convolution layers are divided into two parts by this filter [25]. It splits each convolution layer into a separate layer for filter and another for combining. It helps to reduce the computation and model size. This model contains smaller parameters than others. It is generally applied for mobile and embedded vision applications.

\textbf{DenseNet-169:} This model alleviates the vanishing gradient descent problem. It uses a simple pattern of connectivity that maintains the highest flow of information among layers in backward and forward computation. Inputs from all previous layers pass through feature maps of their own to all subsequent layers. The model is divided into multiple densely connected blocks to reduce the down-sampling. The layers in between these dense blocks accelerate convolution and pooling operations.

\subsection{Evaluation Matrices}
In this section, we discuss several common metrics to evaluate the performance of transfer learning techniques. The matrices that we consider for our classification task are accuracy, precision, recall, F1 score, confusion matrix, and ROC-AUC score. Accuracy is the percentage of the number of correct predictions divided by the total number of predictions. Precision refers to the ratio of correctly predicted images belonging to the positive classes to the total number of images that are predicted as positive by the model. Recall is the ratio of actual positive classes divided by the total number of predicted positive classes. The confusion matrix represents the relationship between the actual class and the predicted class of a classification task. The result of this matrix falls into one of the four classes TP, TN, FP, and FN. The F1 score provides a trade-off between the measures of precision and recall. True positives and true negatives are represented by using TP and TN. FP and FN refer to false positives and false negatives.

\subsection{Data Augmentation}
A large amount of data is needed to achieve the best accuracy of deep learning models. Data augmentation plays a very vital role in training a convolutional neural network. In our dataset, we observe that majority of images are similar which is not advantageous for our models. There are several transformation techniques to perform data augmentation such as rotating, flipping, zooming, and changing brightness. We applied data augmentation in our dataset to increase the number of training images and balance the dataset. Initially, the number of training images for benign and malignant were 1534 and 1284. The total number of training images becomes 5636 after applying augmentation. The below table shows the total number of images used for training, testing, and validation after augmentation.
\begin{table}[htbp]
\caption{Training dataset after augmentation}
\centering
\begin{tabular}{lll}
\toprule
\text{Type} & \text{Before Augmentation} & \text{After Augmentation} \\
\midrule
Benign&1534& 3068\\
Malignant&1284& 2568\\
Total&2818& 5636\\
\bottomrule
\end{tabular}
\end{table}
\subsection{Experimental breakdown}
\begin{itemize}
\item \textbf{Framework:} Keras, a high-level deep learning framework built on tensor flow was used. We used the Google Colab platform for faster execution of model training.
\item \textbf{Dropout:} This regularization technique was used to prevent overfitting.
\item \textbf{Epoch:} Trained each model for 20 epochs with different combinations of hyper-parameters such as learning rate, batch size, and optimizer.
\item \textbf{Learning rate:} We observe model outcomes using two different learning rates such as 0.0001 and 0.00001.
\item \textbf{Batch size:} This is the number of training samples used for one iteration. The experiment tried 32 and 64.
\item \textbf{Optimizer:} It helps with minimizing of loss. Stochastic gradient descent and Adam were experimented.
\item \textbf{Dataset Splitting:} We used 80\% of data for training and 20\% for validation.
\item \textbf{Data augmentation:} We increased the size of the dataset by applying data augmentation such as zooming, flipping, and changing brightness.
\item \textbf{Loss function:} To measure loss we use binary cross entropy loss function.
\item \textbf{Performance metrics:} Existing metrics for the state-of-the-art were used and a new metric was introduced.

\end{itemize}

\begin{table}[htbp]
\caption{Batch size effects}
\centering
\begin{tabular}{llllll}
\toprule
\text{Batch size} & \text{Accuracy} & \text{Precision} &\text{Recall} &\text{F1 Score} &\text{ROC-AUC} \\
\midrule
\textbf {32} & \textbf{0.927}& \textbf{0.91} & \textbf{0.74} & \textbf{0.82} & \textbf{0.844} \\
64&0.935& 0.94 & 0.77 & 0.86 & 0.891 \\
\bottomrule
\end{tabular}
\end{table}

\begin{table}[htbp]
\caption{Learning rate effects}
\centering
\begin{tabular}{llllll}
\toprule
\text{Learning rate} & \text{Accuracy} & \text{Precision} &\text{Recall} &\text{F1 Score} &\text{ROC-AUC} \\
\midrule
\textbf {0.0001} & \textbf{0.935}& \textbf{0.94} & \textbf{0.77} & \textbf{0.86} & \textbf{0.891} \\
0.00001&0.917& 0.89 & 0.73 & 0.80 & 0.823 \\
\bottomrule
\end{tabular}
\end{table}

\begin{table}[htbp]
\caption{Adam vs SGD optimizer.}
\centering
\begin{tabular}{llllll}
\toprule
\text{Optimizer} & \text{Accuracy} & \text{Precision} &\text{Recall} &\text{F1 Score} &\text{ROC-AUC} \\
\midrule
\textbf {Adam} & \textbf{0.935}& \textbf{0.94} & \textbf{0.77} & \textbf{0.86} & \textbf{0.891} \\
SGD&0.904& 0.92 & 0.73 & 0.81 & 0.832 \\
\bottomrule
\end{tabular}
\end{table}

\begin{figure*}
\centering
{\includegraphics[height=4cm, width=0.42\textwidth]{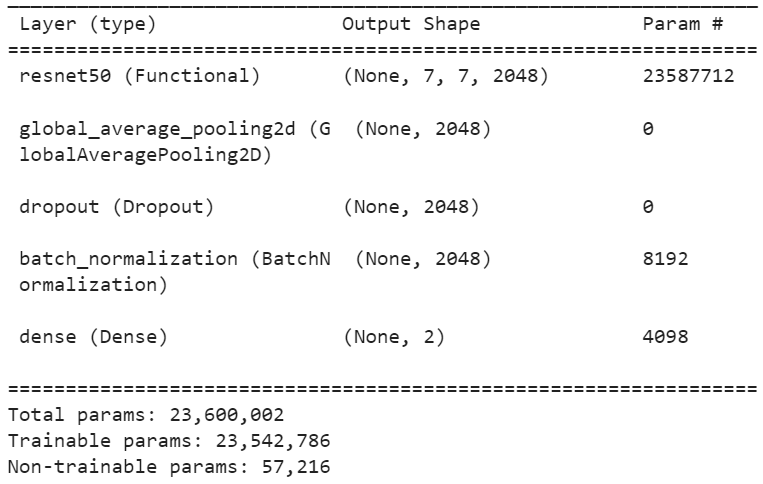}}\label{fig:1}
{\includegraphics[height=4cm, width=0.42\textwidth]{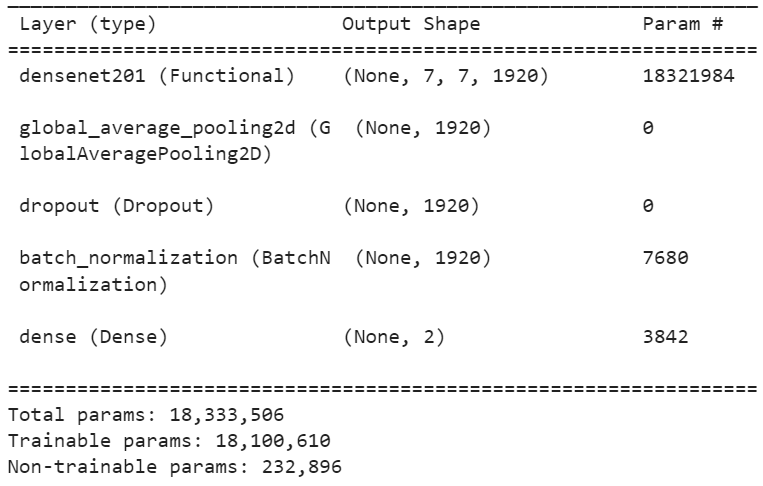}}\label{fig:2}
\caption{ResNet-50 and DenseNet-201 model parameters summary. }
\label{figure}
\end{figure*}

\begin{figure*}
\centering
\subfloat[Loss vs epoch]
{\includegraphics[height=4cm, width=0.42\textwidth]{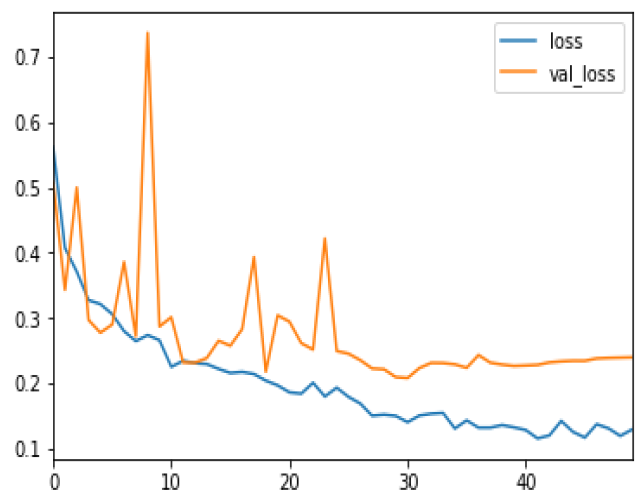}\label{fig:1}}
\subfloat[Accuracy vs epoch]
{\includegraphics[height=4cm, width=0.42\textwidth]{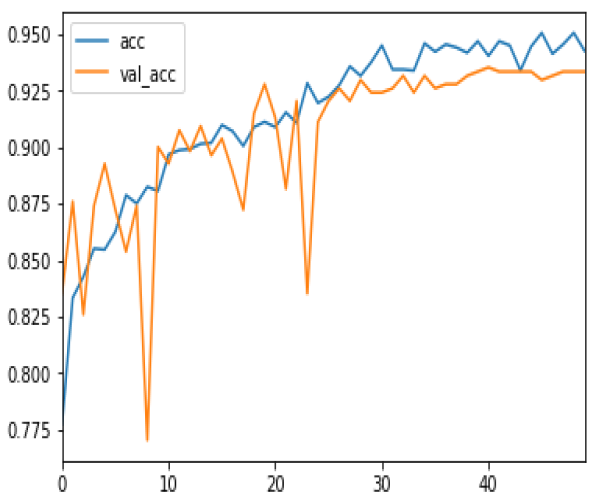}\label{fig:2}}
\caption{Illustration of the loss vs epoch and accuracy vs epoch. }
\label{figure}
\end{figure*}

\begin{figure*}
\centering
\subfloat[Confusion Matrix]
{\includegraphics[height=4cm, width=0.42\textwidth]{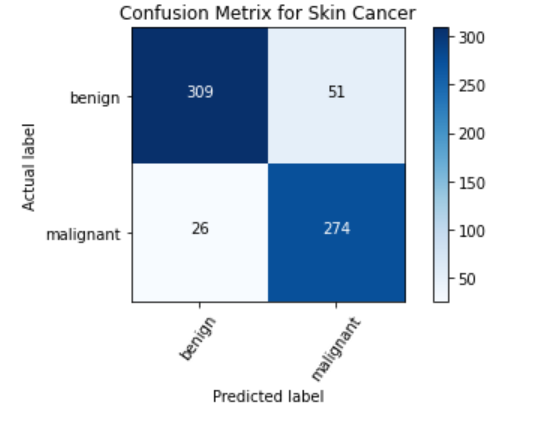}\label{fig:1}}
\subfloat[ROC-AUC curve]
{\includegraphics[height=4cm, width=0.42\textwidth]{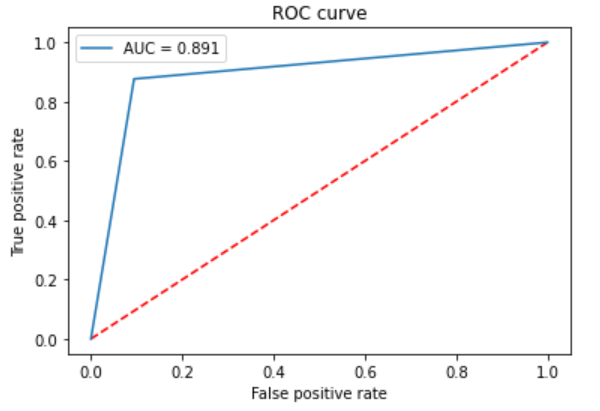}\label{fig:2}}
\caption{Illustration of the confusion matrix and ROC-AUC curve. }
\label{figure}
\end{figure*}
\begin{table}[htbp]
\caption{Comparison of Results.}
\centering
\begin{tabular}{llllll}
\toprule
\text{Model} & \text{Accuracy} & \text{Precision} &\text{Recall} &\text{F1 Score} &\text{ROC-AUC} \\
\midrule
\textbf {ResNet-50} & \textbf{0.935}& \textbf{0.94} & \textbf{0.77} & \textbf{0.86} & \textbf{0.891} \\
MobileNet&0.884& 0.91 & 0.73 & 0.81 & 0.824 \\
InceptionV3&0.884& 0.91 & 0.73 & 0.81 & 0.824 \\
InceptionResnetv2&0.884& 0.91 & 0.73 & 0.81 & 0.824 \\
DenseNet-201&0.884& 0.91 & 0.73 & 0.81 & 0.824 \\
\bottomrule
\end{tabular}
\end{table}

\section{Results and Discussion}
\subsection{Evaluation of network performance}
In this section, we will present our results. This performance for this experiment on a case by case basis is detailed in the table above. Extensive experiments were carried out for each transfer learning model to observe changes in network performance, some of the experiments took longer time to train. Upon completion and comparisons, ResNet-50 gives the best performance. An interesting thing about the performance result is that there is a pattern among the models that shows that each model with batch size 64, learning rate 0.0001, and Adam optimizer provides the best performance. Data augmentation helps to improve performance. We briefly present evaluation matrices for the ResNet-50 model. Fig. 3a, Fig. 3b, Fig. 4a, and Fig. 4b show the graph of loss vs epoch, accuracy vs epoch, confusion matrix, and ROC-AUC curve for the ResNet-50 classifier. 

Table 6 shows the values of Precision, Recall, and F1 score for the prediction on the validation dataset by using the ResNet-50 model. Table 3, Table 4, and Table 5 represent the effect of learning rate, batch size, and optimizer on the model performance for the classification task. We can see that batch size 64 provides slightly better result than 32 for the ResNet-50 model. A learning rate of 0.0001 greatly increases the performance of the model than 0.00001. Moreover, the Adam optimizer accelerates the model performance compared to the stochastic gradient descent (SGD) optimizer. In Fig. 5b, we showed a sample portion of our model prediction on the binary image classification. Table 7 presents the core outcomes of our experiments. 

\begin{figure*}[t]
\centering
\subfloat[Accuracy of different models]
{\includegraphics[height=4cm, width=0.4\textwidth]{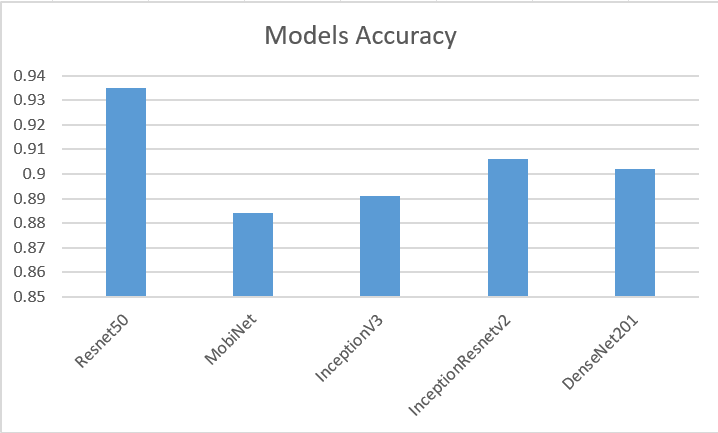}\label{fig:1}}
\subfloat[Sample results of model prediction]
{\includegraphics[height=4cm, width=0.55\textwidth]{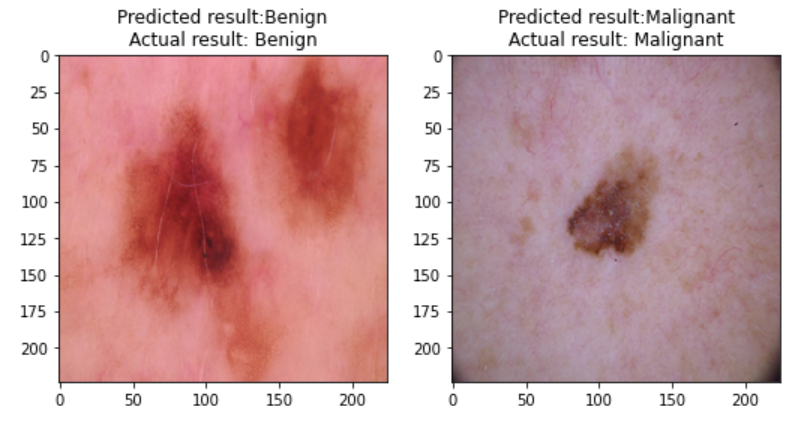}\label{fig:2}}
\caption{Illustration of the accuracy of different models and sample prediction results. }
\label{figure}
\end{figure*}
\begin{table}[htbp]
\caption{Comparison of the proposed model with state-of-the-art techniques.}
\centering
\begin{tabular}{llll}
\toprule
\text{Ref.} & \text{Methodology} & \text{Dataset} &\text{Accuracy} \\
\midrule
{[26]}&InceptionV3, ResNet, VGG19& ISIC & 86.90\% \\
{[27]}&FSM, SVM& Ph2 & 91.90\% \\
{[16]}&MobileNet& HAM10000 & 83.10\% \\
{[11]}&ResNet-50& ISIC & 89.5\% \\
\textbf {Our} & \textbf{ResNet-50}& \textbf{ISIC} & \textbf{94\%} \\
\bottomrule
\end{tabular}
\end{table}
\subsection{Discusion and Results comparison}
Table 6 shows the performance of different transfer learning models on the image classification task. In Table 7, we compare our best-performing model with state-of-the-art methods. Many prior research has been done in this domain. They applied different models on different datasets. Our model achieves comparatively better accuracy than the state-of-the-art methods. Initially, we observe the performance of pre-trained models in different hyperparameter settings. After that, we find out the best-performing settings for each model. Transfer learning techniques greatly reduce the model training time. It provides the opportunity to use pre-trained model weights instead of starting training from scratch.

Another important thing is that the size of the dataset is very small. Data augmentation helps to accelerate the model performance by increasing the amount of training dataset. While the goal of this experimental work is not to compete with the existing state-of-the-art networks, it is equally expedient to compare the best result of this experiment with the former to create room for a measurable future improvement on this network. The below table shows how the network’s performance compares with a few state-of-the-art:

\section{Conclusion and Future Work}
In this work, we have made an experimental study to discover the performance of deep convolutional neural networks on binary classification of skin cancer images. Our experimental results demonstrate that using pre-trained transfer learning models greatly reduces the model training time as well as increase the overall accuracy. Data augmentation accelerates model performance by increasing the amount of training images. Our experimental result outperforms several state-of-the-art works. The best-performing model provides an accuracy of 0.935, precision of 0.94, recall of 0.77, and f1 score of 0.86. This type of computer-assisted diagnosis needs to be implemented in our healthcare centers. It will greatly increase the survival rate by early detecting skin cancer. Furthermore, it makes the costlier treatment more affordable to the poor people.

Many prior researches focused on skin cancer prediction by using different datasets and techniques. It is very essential for healthcare professionals to detect skin cancer at an earlier stage. Automated systems can assist them to detect skin cancer with less effort and time. In this work, we focused on the binary classification of skin cancer images using different transfer learning techniques. In the future, we will investigate the effect of applying transfer learning and data augmentation techniques in the case of multi-class skin cancer image classification tasks with larger datasets.

\section*{References}

\medskip

{
\small

[1]	Kricker A., Armstrong B.K., English D.R. Sun exposure and non-melanocytic skin cancer. Cancer Causes Control. 1994; 5 :367–392. {\it doi: 10.1007/BF01804988}.

[2]	Armstrong B.K., Kricker A. The epidemiology of UV-induced skin cancer.J. Photochem. Photobiol. B. 2001;63:8–18. {\it doi: 10.1016/S1011-1344(01)00198-1}. 

[3]	Christo Ananth and M Julie Therese. A survey on melanoma: Skin cancer through computerized diagnosis. {\it Available at SSRN 3551811, 2020}.

[4] American Cancer Society Cancer Facts and Figures. 2020. [(accessed on 4 April 2024)]. {\it Available online}.

[5]	Silberstein L., Anastasi J. Hematology: Basic Principles and Practice. {\it Elsevier}; Amsterdam, Netherlands: 2017. p. 2408.

[6]	Khalid M Hosny, Mohamed A Kassem, and Mohamed M Foaud. Classification of skin lesions using transfer learning and augmentation with Alex-net. {\it PloS one, 14(5):e0217293}, 2019.

[7]	Divya Gangwani and Pranav Gangwani. Applications of machine learning and artificial intelligence in intelligent transportation system: A review. {\it Applications of Artificial Intelligence and Machine Learning}, pages 203–216, 2021.

[8]	Kazuhisa Matsunaga, Akira Hamada, Akane Minagawa, and Hiroshi Koga. Image classification of melanoma, nevus, and seborrheic keratosis by deep neural network ensemble. arXiv preprint arXiv:1703.03108, 2017.

[9]	Md Ashraful Alam Milton. Automated skin lesion classification using ensemble of deep neural networks in isic 2018: Skin lesion analysis towards melanoma detection challenge. arXiv preprint arXiv:1901.10802, 2019.

[10] A Esteva et al. Dermatologist-level classification of skin cancer with deep neural networks. nat. 542, 115–118, 2017.

[11] Md Ashraful Alam Milton. Automated skin lesion classification using ensemble of deep neural networks in isic 2018: Skin lesion analysis towards melanoma detection challenge. arXiv preprint arXiv:1901.10802, 2019.

[12] Li-Qiang Zhou, Xing-Long Wu, Shu-Yan Huang, Ge-Ge Wu, Hua-Rong Ye, Qi Wei, Ling-Yun Bao, You-Bin Deng, Xing-Rui Li, Xin-Wu Cui, et al. Lymph node metastasis prediction from primary breast cancer us images using deep learning. Radiology, 294(1):19–28, 2020.

[13] Ghaznavi, F., Evans, A., Madabhushi, A., Feldman, M.: Digital imaging in pathol-ogy: whole-slide imaging and beyond. Annual Review of Pathology: Mechanismsof Disease8(2013) 331–359.

[14] Nazeri, Kamyar, Azad Aminpour, and Mehran Ebrahimi. "Two-stage convolutional neural network for breast cancer histology image classification." International Conference Image Analysis and Recognition. Springer, Cham, 2018.

[15] Guo Y., Liu Y., Oerlemans A., Lao S., Wu S., Lew M.S. Deep learning for visual understanding: A review. Neurocomputing. 2016; 187:27–48. doi: 10.1016/j.neucom.2015.09.116. 

[16] Chaturvedi S.S., Gupta K., Prasad P.S. Skin Lesion Analyser: An Efficient Seven-Way Multi-class Skin Cancer ClassificationUsing MobileNet. Adv. Mach. Learn. Technol. Appl. 2020; 1141 :165–176. doi: 10.1007/978-981-15-3383-915.

[17] S. Bassi, A. Gomekar, Deep learning diagnosis of pigmented skin lesions, in: Proceedings of the 10th International Conference on Computing, Communication and Networking Technologies (ICCCNT), IEEE, 2019, pp.1–6. 

[18] Canziani A., Paszke A., Culurciello E. An Analysis of Deep Neural Network Models for Practical Applications. arXiv. 2016 1605.07678 

[19] D. Moldovan, Transfer Learning Based Method for Two-Step Skin Cancer Images Classification, in: 2019 E-Health and Bioengineering Conference, (EHB) 2019 Nov 21, IEEE, pp.1–4. 

[20] Kazuhisa Matsunaga, Akira Hamada, Akane Minagawa, and Hiroshi Koga. Image classification of melanoma, nevus, and seborrheic keratosis by deep neural network ensemble. arXiv preprint arXiv:1703.03108, 2017.

[21] Md Sirajul Islam, Sanjeev Panta. Heart Disease Prediction from Medical Records, 2024. DOI: 10.13140/RG.2.2.28001.52323.
\end{document}